\documentclass[letterpaper, 10 pt, conference]{ieeeconf}  

\IEEEoverridecommandlockouts                              

\usepackage{graphics} 
\usepackage{epsfig} 
\usepackage{times} 
\usepackage{amsmath} 
\usepackage{amssymb}  
\usepackage{mathtools}
\usepackage{graphicx}
\usepackage{tabularx}
\usepackage{algorithm}
\usepackage{algorithmic}
\usepackage{color}
\usepackage{cite}
\usepackage{amsmath,amssymb,amsfonts}
\usepackage{graphicx}
\usepackage{textcomp}
\usepackage{xcolor}
\usepackage{algorithm}
\usepackage{verbatim} 
\usepackage{mathrsfs}
\usepackage{subfigure}
\usepackage{empheq}
\usepackage{lipsum}
\usepackage{physics}
\usepackage{comment}
\usepackage{mathtools, cuted}
\usepackage{bm}
\usepackage[utf8]{inputenc}
\usepackage[english]{babel}
\newcommand{\vg}[1]{\bm{#1}}
\usepackage{hyperref}
\hypersetup{
    colorlinks=true,    
    urlcolor=blue,
}

\renewcommand{\v}[1]{\mathbf{#1}}
\graphicspath{{./figures/}}

\title{\LARGE \bf
SyDeBO: Symbolic-Decision-Embedded Bilevel Optimization for Long-Horizon Manipulation in Dynamic Environments
}

\author{Zhigen Zhao$^{1}$, Ziyi Zhou$^{1}$, Michael Park$^{1, 2}$, and Ye Zhao$^{1}$
\thanks{$^{1}$The authors are with the Laboratory for Intelligent Decision and Autonomous Robots, George W. Woodruff School of Mechanical Engineering, Georgia Institute of Technology, Atlanta, GA 30313, USA.
        {\tt\small \{zhigen.zhao, zhouziyi, michael.j.park, yzhao301\} @gatech.edu}.
}
\thanks{$^{2}$Georgia Tech Research Institute, Atlanta, GA 30318, USA
}
}

\begin{document}
\maketitle
\thispagestyle{empty}
\pagestyle{empty}

\begin{abstract}
%
This study proposes a Task and Motion Planning (TAMP) method with symbolic decisions embedded in a bilevel optimization. This TAMP method exploits the discrete structure of sequential manipulation for long-horizon and versatile tasks in dynamically changing environments. At the symbolic planning level, we propose a scalable decision-making method for long-horizon manipulation tasks using the Planning Domain Definition Language (PDDL) with causal graph decomposition. At the motion planning level, we devise a trajectory optimization (TO) approach based on the Alternating Direction Method of Multipliers (ADMM), suitable for solving constrained, large-scale nonlinear optimization in a distributed manner. Distinct from conventional geometric motion planners, our approach generates highly dynamic manipulation motions by incorporating the full robot and object dynamics. Furthermore, in lieu of a hierarchical planning approach, we solve a holistically integrated bilevel optimization problem involving costs from both the low-level TO and the high-level search. Simulation and experimental results demonstrate dynamic manipulation for long-horizon object sorting tasks in clutter and on a moving conveyor belt.
\end{abstract}

\section{Introduction}

\begin{figure}
    \centering
    \includegraphics[width=0.4\textwidth]{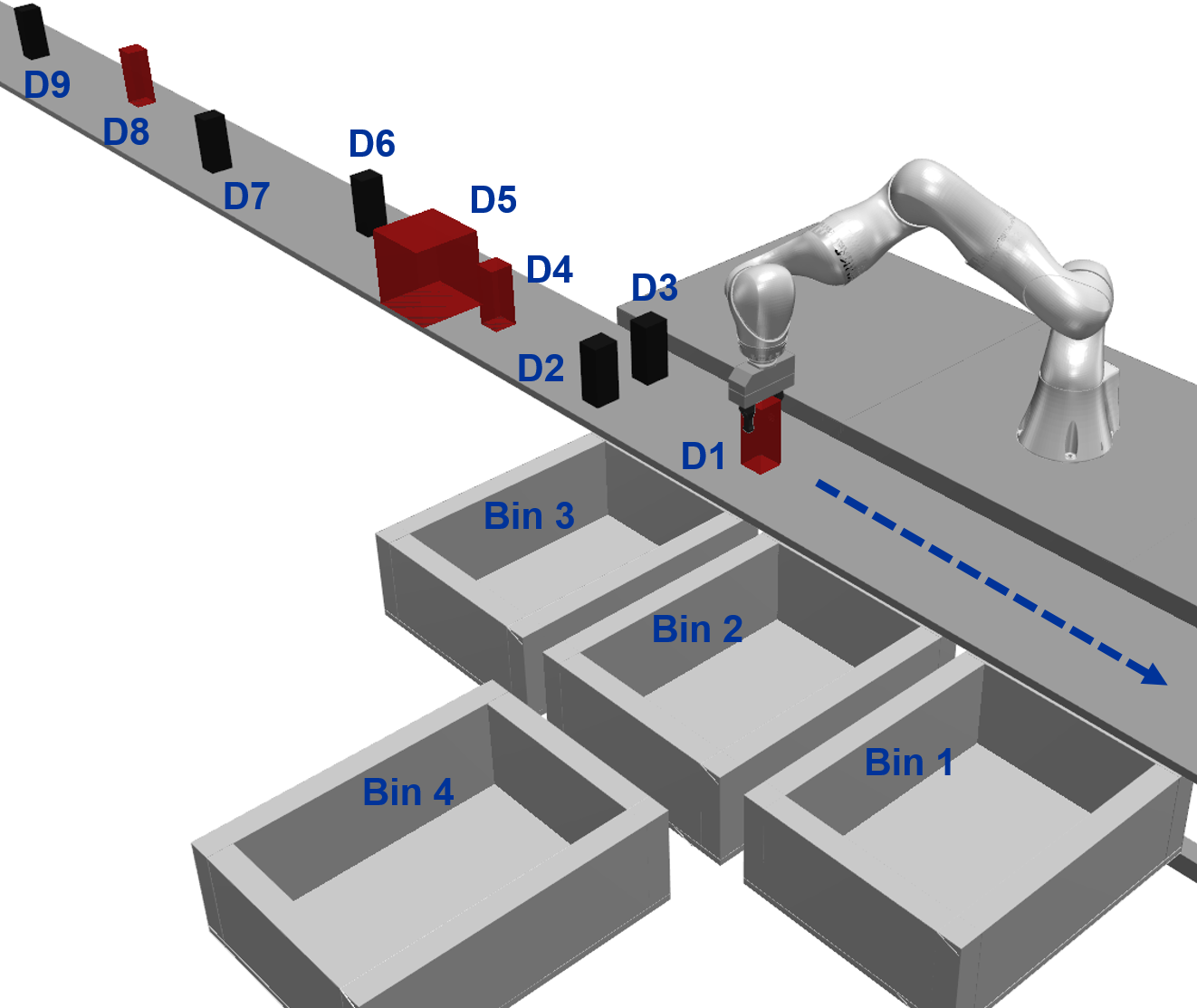}
    \caption{Illustration of a block sorting task in a dynamically changing environment: A set of labeled blocks are placed on a moving conveyor belt adjacent to a set of fixed labeled bins. The robot arm sorts each block into designated bins by executing kino-dynamically feasible trajectories.
    }
    \label{fig:scenario}
    \vspace{-0.1in}
\end{figure}

Long-horizon robot manipulation such as those observed in industrial assembly and logistics (see Figure~\ref{fig:scenario} for a conceptual illustration) often involves hard-coded and repetitive motions. This defect severely limits manipulation task variety and flexibility. Recent attention has been drawn to dynamic manipulation that involves versatile interactions with cluttered environments or complex objects. For instance, how can a robot arm with a two-parallel-jaw gripper manipulate an oversized package by a pushing action or throw a parcel into an unreachable cart? To date, a majority of the existing task and motion planners (TAMPs) lack a formal guarantee of achieving optimal sequential task composition while simultaneously meeting dynamics constraints from the robot manipulators and intricate contact events. This study will take a step towards unifying the high-level task planning and low-level dynamics-consistent trajectory optimization into a coherent TAMP framework for long-horizon manipulation. 



Artificial intelligence (AI) planning approaches have made significant progress in handling symbolic plan search and tasks constraints, manifested in three mainstream paradigms: 1) translating to a Boolean satisfiability problem \cite{kautz2006satplan}; 2) forward state-space search with heuristics \cite{helmert2006fast}; 3) search using planning graph \cite{blum1995fast}. However, such traditional AI planning methods often disregard the low-level physical constraints and dynamics when evaluating the cost and the feasibility of a plan, which poses a challenge to robotics problems often involving complex dynamics and contact physics. To address this problem, our study combines the forward state-space search method with a full-model-based optimization under the framework of bilevel trajectory optimization (TO). To alleviate the computational burden of solving TO within a discrete search, a causal graph decomposition is employed to reduce the size of the discrete planning domain.

State-of-the-art TO methods for contact-rich robotic systems often incorporate intrinsically hybrid contact dynamics either in a hierarchical \cite{carpentier2018multicontact, mastalli2020crocoddyl} or a contact-implicit \cite{posa2014direct,dai2014whole, mordatch2012discovery} fashion, where the TO is formulated either with or without knowing the contact mode sequence a priori. However, existing TO has a limitation in designing robot motions associated with a long sequence of actions. This results from the challenge of designing costs and constraints for a highly nonlinear optimization problem and the difficulty of algorithm convergence over a long trajectory duration. 
This study aims to partially address these challenges by decomposing the long-horizon planning problem into multiple sub-problems for computation efficiency and sequentially solve a combined symbolic planning and TO as bilevel optimization. 

\begin{figure}[t!]
    \centering
    \includegraphics[width=1\linewidth]{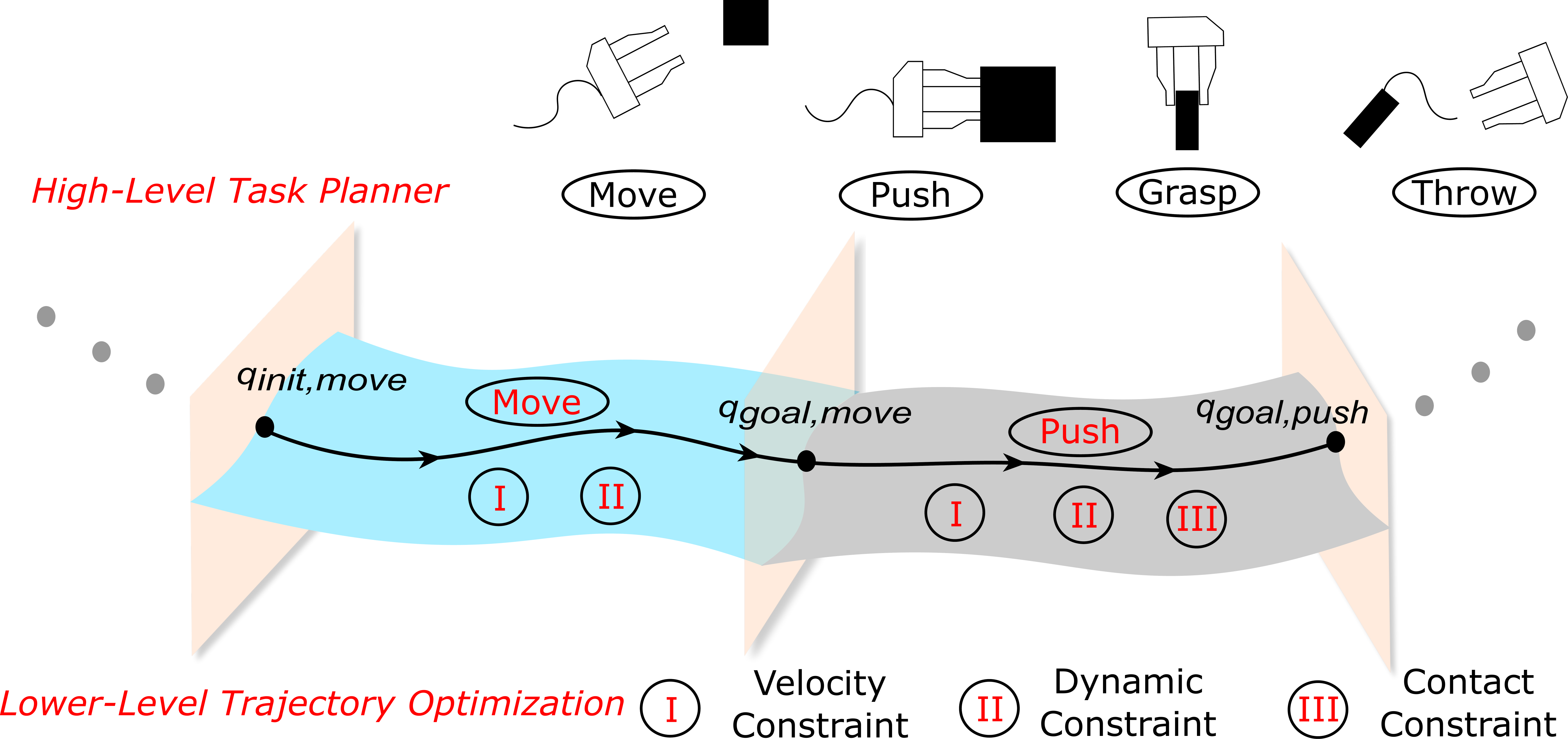}
    \caption{A conceptual illustration of the natural match between the discrete task planner and the low-level distributed trajectory optimization. In the \texttt{move} action without an object in hand, velocity and dynamics constraints are enforced. When the gripper is in the \texttt{push} action, the optimization naturally adds a contact constraint. Note that $q_{\rm goal, move} = q_{\rm init, push}$.
    }
    \label{fig:manifold}
    \vspace{-0.1in}
\end{figure}

The Alternating Direction Method of Multipliers (ADMM) approach \cite{boyd2011distributed} employed in our TO provides a general framework capable of handling various constraints \cite{o2013splitting,sindhwani2017sequential,ZhouAcceleratedDynamics} including contact constraints and manipulation task constraints by introducing multiple ADMM blocks. The discrete states and actions defined in the aforementioned symbolic planning algorithm will be encoded as symbolic constraints in the bilevel optimization and decompose the full optimization into multiple sub-problems. The synthesized high-level symbolic planner will govern the activeness of individual ADMM blocks. Therefore, the distributed nature of our ADMM naturally matches the discrete symbolic search (Figure~\ref{fig:manifold}). 

The proposed symbolic-decision-embedded bilevel optimization (SyDeBO) uses the symbolic decision variables in PDDL to design expressive logic formulas for complex grasping tasks and physical interaction representations.
Our approach exploits the internal structure of the underlying large-scale nonlinear optimization and further split it into multiple sub-problems.
As such, this approach effectively avoids exhaustive exploration of the large state space.

The contributions of this study lie in the following:
\begin{itemize}
\item Propose a causal graph method at the discrete planning domain to identify and decompose independent sub-tasks. The separation of sub-tasks simplifies the entire problem by limiting the actions and predicates to a relevant subset for each sub-problem. 
\item 
Design a holistic bilevel optimization in solving the manipulation sub-task identified above, distinct from the conventional hierarchical planning paradigm. Cost functions from both the discrete actions and TO jointly solve the optimal grasping sequence and motions.
\item Devise an ADMM-based distributed optimization to incorporate various sets of dynamics constraints, which are enforced by the symbolic planner. This distributed structure enables a flexible switching mechanism for activating and deactivating constraint blocks, well suited for being integrated with a discrete symbolic planner.
\end{itemize}


\section{Related Work}
\textit{Task and Motion Planning:} TAMP for dynamic robot manipulation has become an increasingly powerful method to explicitly define symbolic tasks and enable a diverse set of manipulation motions \cite{kaelbling2011hierarchical, hauser2011randomized,  srivastava2014combined, dantam2016incremental, kingston2020informing}. Woodruff and Lynch \cite{woodruff2017planning} proposed a hierarchical planning approach for hybrid manipulation that defined a sequence of motion primitives a priori and stabilized each mode using linear-quadratic-regulator control. However, pure hierarchical planning has the limitation that the high-level mode planner does not take into account the underlying motion plan cost and feasibility. This limitation is often mitigated by embedding sampling-based motion planners as a subroutine to guide the task planning procedure. Garret et. al. proposed a method that incorporated sampling procedures symbolically in PDDL \cite{garrett2017strips}. Along another line of research, Toussaint proposed an optimization-based algorithm that embedded the high-level logic representation into the low-level optimization \cite{toussaint2015logic, toussaint2018differentiable}. The work in \cite{migimatsu2020object} adapted this task and motion planning method to object-centric manipulation in a dynamic environment. However, many existing TAMP works ignore underlying dynamics and physics and only deal with kinematics and geometry constraints. In comparison, our approach takes these ignored constraints into account and falls into the category of kino-dynamics planning.

\textit{Hybrid Planning in Manipulation:} One paradigm conceptually close to our work of incorporating robot dynamics is hybrid Differential Dynamic Programming (DDP), which aims to solve a hybrid optimal control problem combining discrete actions and continuous control inputs. The work in \cite{pajarinen2017hybrid} optimized continuous mixtures of discrete actions. By assigning a pseudo-probability to each discrete action, the control inputs, dynamics model and cost function were expressed in a mixture model suitable for a DDP solver. Along a similar line of research, the authors in \cite{doshi2020hybrid} used an exhaustive search over all hybrid possibilities with the cost computed by an input-constrained DDP. However, both of these two works are limited in a small set of manipulation actions. In our study, more complex manipulation actions are formally defined and sequentially composed via the PDDL-based symbolic planning method.



\textit{Causal Graph in AI Planning:} To address the challenge from the large-scale searching problem for long-horizon manipulation, we propose a causal graph task decomposition. Causal graphs have been used in AI planning domain to  construct local subproblems for a vertex and estimate the search heuristics by traversing the pruned causal graph and looking for paths in the corresponding domain transition graphs \cite{helmert2004planning, helmert2006fast}. However, these methods are primarily used for purely symbolic planning problems, where the search algorithm has full knowledge of the path costs on the domain transition graphs. In our method, a causal graph is used to identify and decompose the symbolic subtasks globally. This subtask decomposition enables our algorithm to scale up for manipulating a series of objects in complex scenarios. 

\textit{Distributed Trajectory Optimization:} ADMM has gained increasing attention in the robotics arena for solving parallel, large-scale motion planning problems. As a special case of Douglas-Rachford splitting methods \cite{eckstein1992douglas}, the classical ADMM was formulated to solve an optimization where the cost function is separable into two sub-problems along with a coupled linear constraint \cite{boyd2011distributed}. 
ADMM has been further explored in \cite{o2013splitting,o2016conic} to solve constrained optimization problems with box and cone constraints. Although formally provable convergence for nonconvex problems can only be guaranteed under specific assumptions \cite{hong2016convergence}, ADMM is powerful in practice and has been widely studied for nonlinear robotic problems  \cite{sindhwani2017sequential, ZhouAcceleratedDynamics}. Our previous work \cite{ZhouAcceleratedDynamics, wijayarathne2020simultaneous} proposed a framework embedding DDP as a sub-problem to solve rigid body dynamics. Inspired by these works above, this study formulates a bilevel optimization that combines an ADMM-based TO with a high-level multi-stage search.

\textit{Bilevel Optimization}:  As an optimization problem embedding another optimization problem as a constraint, bilevel optimization gained significent attention within the mathematical
programming community \cite{colson2007overview}. In legged locomotion, bilevel optimization has been widely adopted to optimize the switching times for hybrid systems \cite{xu2004optimal,egerstedt2003optimal,farshidian2017efficient}. In \cite{carius2018trajectory} the computation of a contact model is formulated as a bottom-level optimization inside a single-shooting TO. The work in \cite{tang2020enhancing} decomposes the time-optimal planning for UAV into spatial and temporal layers. Along another line of AI research, \cite{stouraitis2020online} proposes a bilevel structure to combine continuous optimization with discrete graph-search for collaborative manipulation, which is closest to the framework presented in this letter. This bilevel formulation not only mitigates the poor convergence of solving the whole optimization problem, but also holistically reasons about variables from each level. Our bilevel optimization implements the low-level TO through ADMM, the distributed structure of which fits well into the high-level discrete search.


\section{Problem formulation}
\subsection{Causal Graph Task Decomposition}
Task and motion planning for sequential manipulation with multiple objects often suffers from heavy computational burden. Exploring each discrete state at the task planning level will require a trajectory cost evaluation of the underlying motion planner. 
However, the number of symbolic states grows exponentially with that of manipulated objects within the planning domain. Therefore, the number of trajectory cost evaluations becomes extremely large and results in computational intractability of the combined TAMP problem.

To mitigate the exponential growth of the symbolic state space, we decompose the entire symbolic task into independent sub-tasks by analyzing the causal graph of the symbolic planning domain. A causal graph is constructed similarly to the definition by Helmert \cite{helmert2006fast}.

\begin{figure}
    \centering
    \includegraphics[width=1\linewidth]{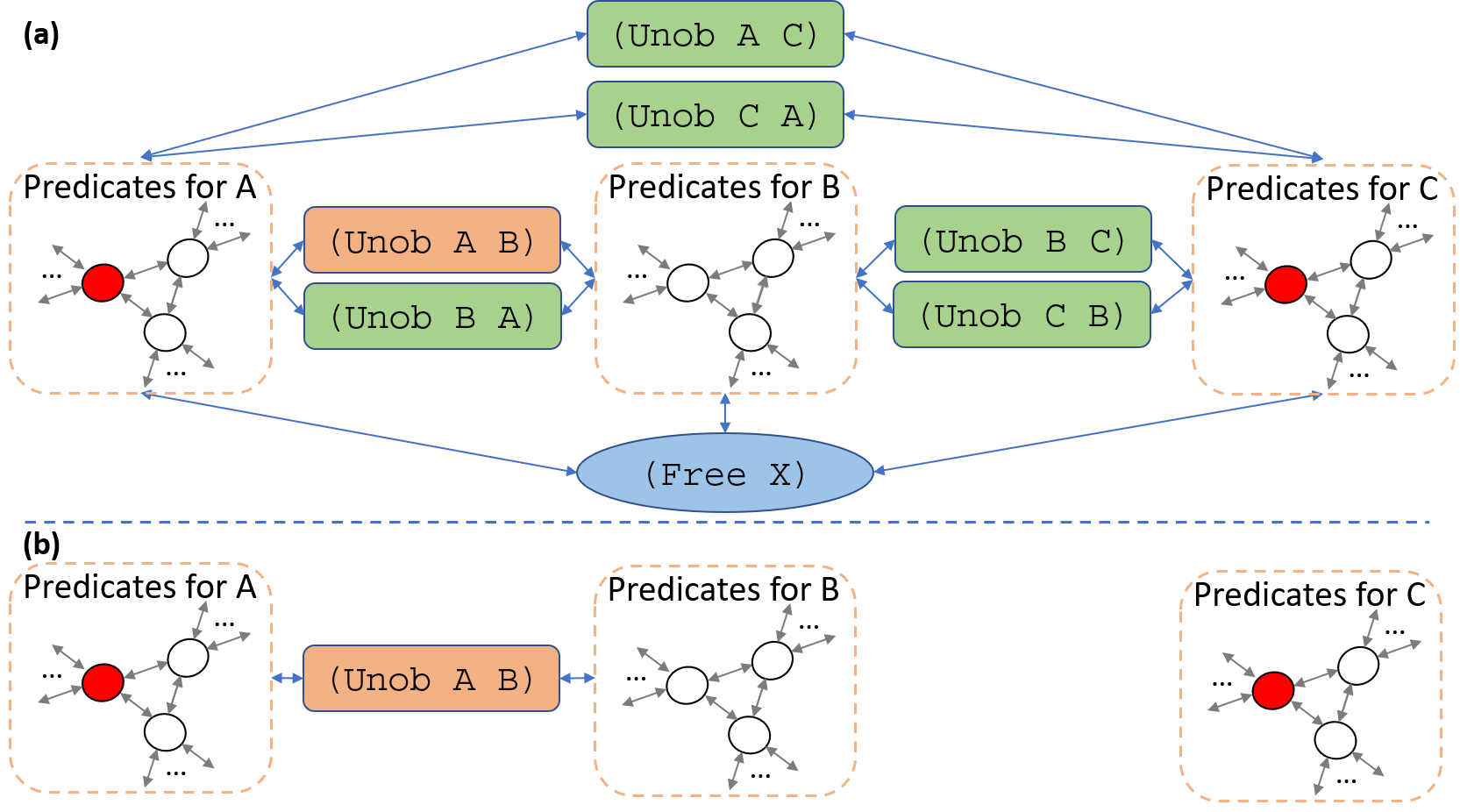}
    \caption{A causal graph illustration for multi-object manipulation task decomposition. The light green indicates a true \texttt{unobstructed} predicate, while the light orange represents a false one. The pruned causal graph in (b) is obtained after decomposing the original causal graph (a). In each dashed block, the sub-graph for each object has the same structure because each object has the same set of predicates. The red circle represents a sub-goal. The abbreviation \texttt{(Unob A B)} stands for \texttt{(Unobstructed A B)}. Symbols A, B, and C index three objects while symbol X denotes the robot.}
    \label{fig:causal_graph}
    \vspace{-0.1in}
\end{figure}

Constructing the causal graph allows us to decouple the unrelated sub-problems of the planning domain by pruning the entire graph into disconnected components, each of which can be solved independently. In the object manipulation example, Figure~\ref{fig:causal_graph} shows two types of vertices eliminated from the graph: \texttt{(free X)} and any \texttt{(unobstructed Y Z)} evaluated to be true. By pruning the \texttt{(free X)} vertex, we relaxed the constraint that the robot arm can only be either empty or holding one object at the same time. For the purpose of task separation, this simplification does not lead to a significant loss of information because the \texttt{(free X}) constraint is still followed when each of the sub-tasks is solved. The \texttt{(unobstructed A B)} predicate indicates whether object B is obstructing the robot arm's reach to object A. When B is not obstructing A and is not in the chain of objects that are obstructing A, any robot manipulation of object A can be solved irrelevant to object B. Therefore, if \texttt{(unobstructed A B)} is true, it can be pruned from the causal graph in order to explicitly decouple the predicates related to object A and B. In the example depicted in Figure \ref{fig:causal_graph}, \texttt{(unobstructed A B)} is false while all other \texttt{unobstructed} predicates are true. The resulting pruned causal graph contains two disconnected components that contain sub-goal predicates. This indicates that the full discrete planning can be divided into two independent subtasks. 

\subsection{Symbolic-Decision-Embedded Bilevel Optimization}
In order to solve for the lowest cost trajectory that achieves the symbolic sub-goal specified by PDDL and causal graph, the TAMP problem is formulated as a bilevel optimization framework, inspired by Toussaint's logic-geometric programming \cite{toussaint2015logic} and Stouraitis's bilevel optimization \cite{stouraitis2020online}. Given initial and final symbolic states $\v s_0, \ \v s_K$ from the decomposed sub-task, the optimization will solve a sequence of discrete actions $\v A= (\v a_{1},\ldots,\v a_{K-1})$ resulting in a sequence of symbolic states $\v S=(\v s_1,\ldots,\v s_K)$, such that the total cost function $\mathcal{J}$ is minimized. Meanwhile, between each symbolic state pair $(\v s_k, \v s_{k+1})$, a local state trajectory segment $\v X_k = (\v x_1,\ldots,\v x_{N_k})$ and a control trajectory segment $\v U_k = (\v u_1,\ldots,\v u_{N_k-1})$ are optimized and the associated costs are incorporated into the high-level graph-search. $K$ denotes the number of discrete symbolic states and $N_k$ represents the number of knot points for the $k^{\rm th}$ trajectory segment. This bilevel optimization is formulated as:
 \begin{subequations}\label{original_formulation}
    \begin{align}\nonumber
      &\underset{\v S,\v A}{\min}  &&\sum_{k=0}^{K-1} \ \Big[\mathcal{J}_{\rm path}(\v s_k, \v a_k) + \mathcal{J}_{\rm discrete}(\v s_{k}, \v a_k)\Big] + \mathcal{J}_{\rm goal}(\v s_K) \\
     & \hspace{0.05in} \text{s.t.}  && \v s_0 = \v s_{\rm init}, \ \v s_K = \v s_{\rm goal}, \; \v a_k \in \mathcal{A}(\v s_k), \\\label{eq:switch_constraint} &&& h_{\rm switch}(\v s_k, \v a_k) = 0, \ g_{\rm switch}(\v s_k, \v a_k) \leq 0,
     \\\label{eq:traj_opt_succ}
            &&& \v s_{k+1} \in \begin{cases}
            \underset{\v X_k,\v U_k}{\min} &\sum_{i=0}^{N_k-1} \mathcal{L}_{\rm path}(\v x_i, \v u_i, \v a_k) \\& + \mathcal{L}_{\rm goal}(\v x_N, \v a_k)\\
            \hspace{0.09in} \text{s.t.} & \v x_{i+1}=\v f_k(\v x_i,\v u_i) \\& \v x_0 = \mathbb{X}_{\rm init}(\v s_{k})\\
        & \v x_N = \mathbb{X}_{\rm goal}(\v s_k, \v a_k)\\
            & \v X_k \in \mathcal{X}_{\v a_k} , \ \v U_k \in \mathcal{U}_{\v a_k}\\
            &\forall i \in [0, N_k-1]
            \end{cases}
    \end{align}
\end{subequations}
where the path cost $\mathcal{J}_{\rm path}$ is composed of the local cost from lower-level trajectory optimization, i.e., $\mathcal{L}_{\rm path}$ and $\mathcal{L}_{\rm goal}$.
$(h, g)_{\rm switch}$ in Eq.~(\ref{eq:switch_constraint}) denotes switch constraints given a symbolic state $\v s_k$ or an action $\v a_k$ that induces a state transition. $\v a_k \in \mathcal{A}(\v s_k)$ indicates the set of all possible actions associated with a symbolic state $\v s_k$. $\v a_k$ is imposed over a specific trajectory segment in (\ref{eq:traj_opt_succ}), which also governs the activeness of different sets of constraints corresponding to specific actions such as moving, pushing or throwing. Therefore, the symbolic-level transition is achieved through the continuous lower-level optimization (\ref{eq:traj_opt_succ}). 

The discrete cost $\mathcal{J}_{\rm{discrete}}$ is defined at the discrete action level to encourage high-priority actions. Let us take the conveyor belt as an example: a lower discrete cost is applied to encourage grasping from the top of an object, to avoid a collision with other objects when grasped from the side. If picking up one object has a higher priority over other objects, a higher discrete cost will be designed to penalize picking up other objects. To reflect the grasping priority, each object is assigned a base discrete cost $\mathcal{J}_{\rm object}$. At each symbolic node, the discrete costs for all unprocessed objects are scaled by a factor parameter $\alpha > 1$. Therefore, the sum of all object costs is $\Sigma_{p=0}^{P-1} \alpha^p\mathcal{J}_{\rm object, p}$, where $P$ is the number of objects. This cost is minimized to generate an optimal grasping sequence taking into account the object priority costs.

For the TO problem in (\ref{eq:traj_opt_succ}), the associated action $\v a_k$ and current state $\v s_k$ determine the TO initial and goal states. The system dynamics constraint associated with the $k^{\rm th}$ discrete mode is expressed as $\v f_k(\v x_i, \v u_i)$. We define feasible sets $\mathcal{X}_{\v a_k}$ and $\mathcal{U}_{\v a_k}$ to represent additional constraints such as the joint, torque and contact constraints subject to action $\v a_K$. Note that the contact force is part of the state $\v x$ for contact modes with a grasped object. We solve the TO problem with ADMM. In particular, our ADMM consists of two blocks: (i) an unconstrained Differential Dynamic Programming (DDP) block, (ii) a constraint block handling constraints such as $\mathcal{X}$ and $\mathcal{U}$. More details will be elaborated in Sec.~\ref{title:distributed_TO}.

\subsection{Symbolic Multi-Stage Search} To solve the bilevel optimization problem defined in Sec. III-B, we extend greedy-best-first search to employ trajectory optimization as a subroutine to estimate path cost of each edge on the tree, where symbolic state $s_{k}$ activates and deactivates corresponding ADMM blocks to apply appropriate constraints. Each node $\mathcal{N}$ of the search tree contains the information of its parent $\mathcal{P}$, children $\mathcal{C}_{i} \in \mathcal{C}$, total cost-so-far, and the symbolic action leading up to the node, and the trajectory optimization solved on the edge between the node and its parent. For each node, we first use a geometric query module to compute the desired robot end-effector pose required by the action and check the kinematic feasibility of the action by inverse kinematics. If the action is feasible, we compute the trajectory cost from its parent using DDP and update its total path cost. Then a discrete exploration of search tree is done to find any symbolic goal node within a predefined number of discrete actions. Nodes to be explored are ranked in a priority queue $\mathcal{F}$. When selecting a new node to visit, priority is given to the nodes with symbolic goal node in its sub-tree, and with the lowest total cost from root. After a feasible solution is found using DDP, ADMM is used to refine the trajectories in order to comply with the kino-dynamic constraints. If a feasible trajectory cannot be found by ADMM, the second best solution will be found via DDP and refined using ADMM again. This process starts from the root node $\mathcal{R}$ with an initial system state, and is repeated until either an ADMM solution is generated, or the tree is fully explored. A pseudo code is depicted in Algorithm~1.

\subsection{Symbolic Action Representation}
\label{sec:actions}
Our object sorting tasks define five types of symbolic manipulation actions including \texttt{Grasp(X, Y, Z)}, \texttt{Move(X, Z)}, \texttt{Release(X, Y, Z)}, \texttt{Push(X, Y, Z)}, and \texttt{Throw(X, Y, Z)}. Details of other actions, preconditions, and effects are shown in Table~I.
%
%
\texttt{Grasp (X, Y, Z)} action allows a robot X to grasp an object Y on a surface Z, either from the top or the side. The preconditions of this grasping action are threefold: (i) the end-effector is not holding any object, (ii) the end-effector is in a ready position to grasp the object Y, and (iii) the object Y is on the surface Z. The robot holds the target object from the top or the side as the outcome of this action. The effect is that the robot is holding the target object from the front. Grasping from the front of the target object frequently gives lower control costs and is more robust against variations in the timing of grasp execution. 

\texttt{Move}$(X, Z)$ action allows a robot X to move to a location Z. The preconditions of the move actions are independent from the grasp and the release actions, meaning that the gripper may or may not be holding an object. The robot X is located at the position Z as the outcome of this action.

\texttt{Release}$(X, Y, Z)$ action allows a robot X to place a object Y at a location Z. The preconditions of this action include that the end-effector is moved to the drop-off position while holding the object. As the outcome, the robot no longer holds the object, and the object is placed at the location Z.

\begin{figure}
    \centering
    \includegraphics[width=3.4in]{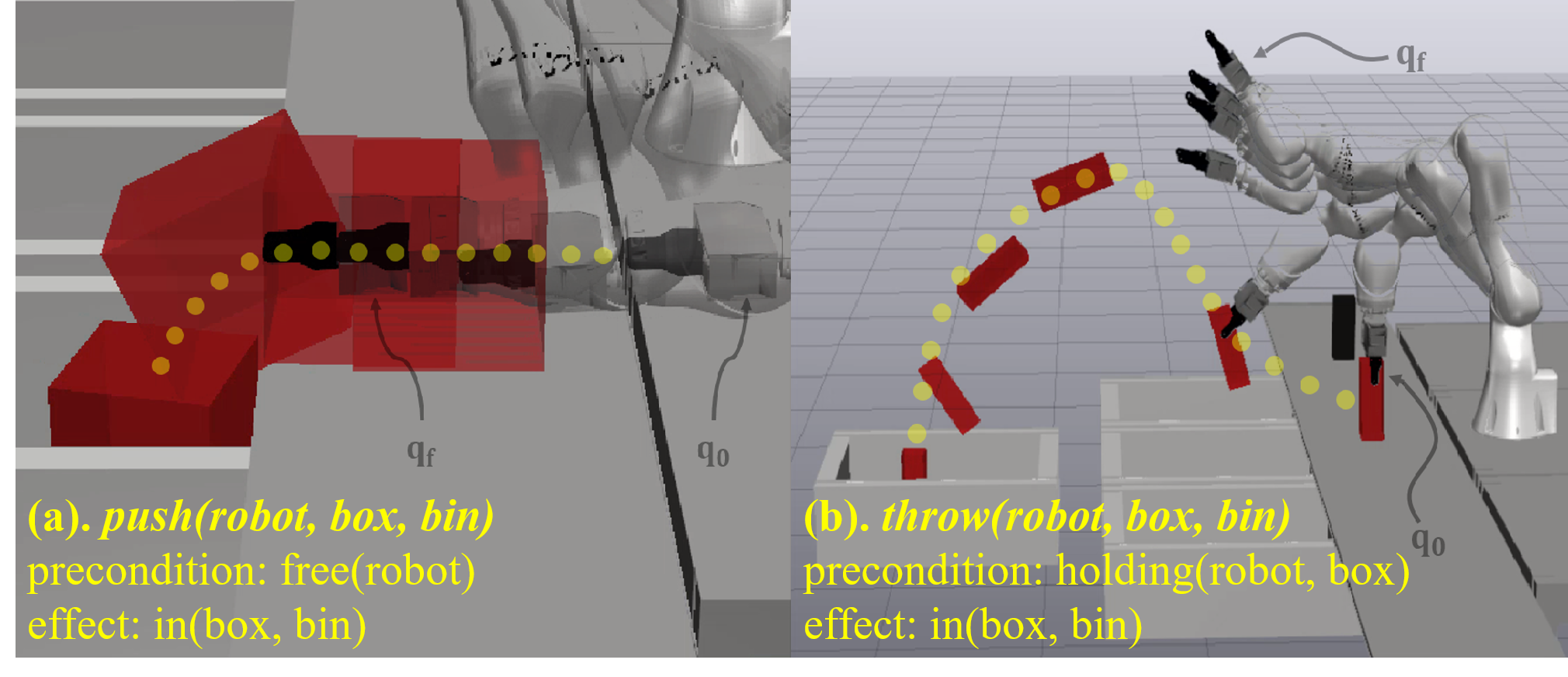}
    \caption{Snapshot of two symbolic actions introduced in Section~\ref{sec:actions}: \texttt{Push} in (a) and \texttt{Throw} in (b). The initial and final states for each action are labeled as $(q_0, q_f)$. Object trajectories are marked with yellow dots.}
    \label{fig:push-throw-actions}
    \vspace{-0.1in}
\end{figure}

\texttt{Push}$(X, Y, Z)$ action allows a robot X to push an object Y to a location Z (see Figure \ref{fig:push-throw-actions}(a)). The preconditions of this action are that the end-effector is moved to a ready position for pushing without holding any objects. The effect of the action is that the object Y is placed at the location Z. 

\texttt{Throw}$(X, Y, Z)$ action allows a robot X to throw an object Y to a location Z. The preconditions are that the end-effector is holding object Y, and that the robot is moved to a position ready for throwing. The effect of the action is that object Y is at location Z. 
After the gripper release, the object follows a free fall motion (see Figure \ref{fig:push-throw-actions}(b)).


\begin{table*}[h]
\caption{Symbolic Actions, Preconditions, and Effects Imposed on Trajectory Optimization}
\vspace{-0.1in}
\label{actions_table}
\begin{center}
\begin{tabular}{|c||c||c|}
    \hline
    Actions & Preconditions & Effects \\
    \hline
    \texttt{Grasp-Top (X, Y, Z)}   & (\texttt{Free X}) (\texttt{At X, Y}) (\texttt{On Y, Z}) & (\texttt{Holding X, Y}) (\texttt{not} (\texttt{Free X})) (\texttt{not} (\texttt{On Y, Z} )) \\
    \hline
    \texttt{Release (X, Y, Z)} & (\texttt{Holding X, Y}) (\texttt{At X, Z}) & (\texttt{Free X}) (\texttt{In Y, Z} ) \\
    \hline
    \texttt{Move-Top (X, Z)} / \texttt{Move-Side (X, Z)}  & N/A & (\texttt{At X, Z}) \\
    \hline
    \texttt{Push (X, Y, Z)} & (\texttt{Free X}) & (\texttt{In Y, Z})\\
    \hline
    \texttt{Throw (X, Y, Z)} & (\texttt{Holding X, Y}) & (\texttt{In Y, Z})\\
    \hline
\end{tabular}
\vspace{-0.2in}
\end{center}
\end{table*}


    
    

\begin{algorithm}[t]
\caption{Multi-stage Search}
\label{alg:ADMM-A-star}
\textbf{Input: } Root node $\mathcal{R}$
\begin{algorithmic}
\STATE $\mathcal{F} \gets \rm{empty\ priority\ queue}$
\STATE $\mathcal{F}\rm{.push(\mathcal{R})}$
\WHILE{$\mathcal{F} \neq \emptyset$}
    \STATE $\mathcal{N} \gets \mathcal{F}\rm{.pop()}$
    \STATE $\mathcal{P} \gets \mathcal{N}\rm{.parent}$
    \IF{$\rm{inverseKinematics(\mathcal{P}, \mathcal{N})}\bf{\ is\ not\ feasible}$}
    \STATE $\mathcal{J}_{\rm path} \gets \infty$
    \STATE $\bf{continue}$
    \ENDIF
    \STATE $\rm{\mathcal{J}_{path}} \gets \rm{DDP(}\mathcal{P}, \mathcal{N}\rm{)}$
    \STATE $\rm{\mathcal{N}.cost} \gets \rm{\mathcal{P}.cost} + \rm{\mathcal{J}_{\rm path}} + \rm{\mathcal{J}_{\rm discrete}}$
    \IF{$\rm{isGoal(\mathcal{N})}$}
        \IF{$\rm{ADMM(}\mathcal{R}, \mathcal{N}\rm{)} \bf{\ is\ feasible}$}
            \STATE return $\mathcal{N}$
        \ENDIF
    \ELSE
    \STATE $\rm{discreteExploration(\mathcal{N})}$
    \FOR{$\mathcal{C}_{i}\ \bf{in}\ \mathcal{C}$}
    \STATE $\rm{\mathcal{F}.push(\mathcal{C}_{i})}$
    \ENDFOR
    \ENDIF

\ENDWHILE
\end{algorithmic}
\end{algorithm}

\section{Distributed Trajectory Optimization}\label{title:distributed_TO}
To enable discrete transitions in symbolic-level search for manipulation tasks, a distributed trajectory optimization -- Alternating Direction Method of Multipliers (ADMM) -- is solved to generate kino-dynamics consistent motions at the low-level. The high-level manipulation actions will govern the activeness of different ADMM blocks in the optimization.
\subsection{Operator Splitting via ADMM}
We first review the basics of the ADMM approach. Consider a general two-block optimization problem with consensus constraints:
\begin{equation}
\begin{aligned}
    & \underset{\v x, \v z}{\text{min}} \ f(\v x) + g(\v z) \quad \text{s.t.} \ \v x = \v z
\end{aligned}
\end{equation}
where two sets of variables $\v x$ and $\v z$ construct a separable cost function and a linear constraint. $f$ and $g$ can be non-smooth or encode admissible sets using indicator functions \cite{o2016conic}. The ADMM algorithm splits the original problem into two blocks and iteratively updates the primal and dual variables as below until the convergence under mild conditions \cite{boyd2011distributed}.
\begin{subequations}
\begin{align}\label{ADMM_a}
    & \v x^{p+1} = \underset{\v x}{\arg\min} \ (f(\v x)+\frac{\rho}{2}\|\v x - \v z^p + \v w^p\|^2) \\\label{ADMM_b}
    & \v z^{p+1} = \underset{\v z}{\arg\min} \ (g(\v z)+\frac{\rho}{2}\|\v x^{p+1} - \v z + \v w^p\|^2) \\\label{ADMM_c}
    & \v w^{p+1} = \v w^p + \v x^{p+1} - \v z^{p+1} 
\end{align}
\end{subequations}
where $p$ denotes the ADMM iteration, $\v w$ is the scaled dual variable and $\rho$ is the penalty parameter.

Assuming that $g$ is an indicator function $I_{\mathcal{B}}$ of a closed convex set $\mathcal{B}$ 
\begin{equation}
    g(\v z)=I_{\mathcal{B}}(\v z)=\begin{cases}
    0, \ \v z\in \mathcal{B} \\
    +\infty, \ \text{otherwise}
    \end{cases}
\end{equation}
we can rewrite Eq.~(\ref{ADMM_b}) as
\begin{subequations}
\begin{align}\nonumber
    \v z^{p+1} = \underset{\v z \in \mathcal{C}}{\arg\min} \ (\frac{\rho}{2}\|\v x^{p+1} - \v z + \v w^p\|^2)
    = \Pi_{\mathcal{B}} \ (\v x^{p+1} + \v w^p)
\end{align}
\end{subequations}
where $\Pi_{\mathcal{B}}(\cdot)$ is a projection operator that projects the input argument onto an admissible set $\mathcal{B}$.

\subsection{ADMM Block Design for Manipulation}
To generate dynamically feasible trajectories given high-level manipulation actions, a set of ADMM blocks are constructed to match the manipulation predicates. In this section, we formulate the low-level ADMM-based trajectory optimizer for versatile manipulation.

As described in the previous subsection, the global optimization problem in Eq.~(\ref{original_formulation}) is composed of a high-level symbolic planner and a low-level trajectory optimizer. The low-level optimization problem is formulated as
\begin{subequations}\label{low-level OC}
    \begin{align}
      \min_{\v X,\v U} \quad &  \sum_{i=0}^{N} \mathcal{L}(\v x_i, \v u_i,\v a)
      \\
     \text{subject to} \hspace{0.08in} & \v x_0 = \v x_{\rm init}, \; \v x_N = \v x_{\rm goal},\\\label{RK4}
           & \v x_{i+1}=\v f(\v x_i,\v u_i),\\ \label{torque_joint_limit_lowlevel}
             &\v X \in \mathcal{X}_{\v a} , \ \v U \in \mathcal{U}_{\v a},\\
            &\v a \in \mathcal{A}
\end{align}
\end{subequations}
where the action $\v a\in \mathcal{A}$ is sent from the high-level symbolic planner. Here we ignore the subscript $k$ for simplicity. To save space, we use $\mathcal{L}$ to denote the low-level cost function comprising $\mathcal{L}_{\rm path}$ and $\mathcal{L}_{\rm goal}$ defined in Eq.~(\ref{original_formulation}). The design of the cost function $\mathcal{L}$ and additional constraints such as joint limits, torque limits and friction cone constraints, vary when different actions $\v a$ are active. The state is defined as $\v x=(\v q,\dot{\v q}, \vg \lambda)^T$, where $\v q$ and $ \dot{\v q}$ are the joint configuration and velocity vectors, respectively. When the gripper manipulates an object, the state $\v x$ includes the object states and $\vg \lambda$ represents the stacked contact force. Otherwise, the object state and the contact force are dropped. 
The dimension of control input $\v u \in \mathbb{R}^m$ is always the same. 
The dynamics constraint $\v f$ represents rigid body dynamics and is numerically integrated by a $4{\rm th}$-order Runge-Kutta method. For a contact-involved action, a fully actuated manipulator with $n$ DoFs and a passive object with 6 DoFs will be modeled as the following
\begin{algorithm}\label{pseudo:DDO-ADMM}
  \caption{DDP-ADMM solver}
    \textbf{Input: } Parent node $\mathcal{P}$, Current node $\mathcal{N}$\\\vspace{-0.17in}
  \begin{algorithmic}
  \IF{$\rm \mathcal{N}.stage=1$}
   \STATE $\vg \phi \gets \vg \phi_{\rm random}^{0}$
   \ELSIF{$\rm \mathcal{N}.stage=2$}
   \STATE $\vg \phi \gets \vg \phi_{\mathcal{N}_1}^{0}$
   \ENDIF
   
   \STATE $\bar{\vg \phi} \gets \bar{\vg \phi}^{0}, \v w_j \gets \v w_j^0,\v w_u \gets \v w_u^0$
   
    \REPEAT
    \IF {$\rm \mathcal{N}.stage=1$}
    \STATE $\vg \rho=0$
    \ENDIF
    \STATE $\vg \phi \gets \text{DDP}\;(\vg \phi,\bar{\v x} - \v w_{j},\bar{\v u}-\v w_{u},\vg \rho)$
    \STATE $\bar{\vg \phi} \gets \text{Projection}\;(\v x+\v w_j,\v u+\v w_u,\rm \mathcal{N}.limits)$
    \STATE $\v w_j \gets \v w_j + \v x - \bar{\v x}$
    \STATE $\v w_u \gets \v w_u + \v u - \bar{\v u}$
    \UNTIL{$\rm{stopping\ criterion\ is\ satisfied}$} \OR{$\rm \mathcal{N}.stage=1$}
    \STATE $\mathcal{J}_{\rm path} \gets \mathcal{L}(\vg \phi,\v a)$
    \RETURN {$\mathcal{J}_{\rm path}, \vg \phi$}
  \end{algorithmic}
\end{algorithm}
%
\begin{equation}
\begin{aligned}\label{object_arm_dynamics}
    \underbrace{\begin{bmatrix}
    \v M_o(\mathbf{q}_o) & \v 0_{6 \times 6}\\
    \v 0_{n \times n} & \v M_r(\mathbf{q}_r)
    \end{bmatrix}}_{\v M(\v q)}\underbrace{\begin{bmatrix}
    \ddot{\mathbf{q}}_o\\
    \ddot{\mathbf{q}}_r
    \end{bmatrix}}_{\ddot{\v q}}& +\underbrace{\begin{bmatrix}
    \v C_o(\mathbf{q}_o,\dot{\mathbf{q}}_o)\\
    \v C_r(\mathbf{q}_r,\dot{\mathbf{q}}_r)
    \end{bmatrix}}_{\v C(\v q, \dot{\v q})}=
    \underbrace{\begin{bmatrix}
    \v 0_{6 \times m}\\
     \v I_{n \times m}
    \end{bmatrix}}_{\v B}\vg \tau\\
    & + \v J_{oc}(\mathbf{q})^T \vg \lambda + \begin{bmatrix}
    \v F_{\rm ext}\\
    \v 0_{n \times 1}
    \end{bmatrix}
\end{aligned}
\end{equation}
where the subscripts $o$ and $r$ represent the object and robot arm, respectively. $\v M \in \mathbb{R}^{(n+6)\times (n+6)}$ is the mass matrix; $\v C \in \mathbb{R}^{n+6}$ is the sum of centrifugal, gravitational, and Coriolis forces; $\v B \in \mathbb{R}^{(n+6) \times m}$ is the selection matrix for control inputs, which consists of a zero matrix for the object and an identity matrix for the  manipulator; $\v F_{\rm ext} \in \mathbb{R}^6$ denotes the external force applied on the object, such as the contact force exerted by the table in the pushing action. We define $\vg \phi(\v q)$ as the signed distances between contact points and the object surface in the object's frame. Then the stacked contact Jacobian matrix is expressed as $\v J_{oc}(\v q) = \pdv{\vg \phi(\v q)}{\v q}$. Since the contact mode is known a priori in (\ref{low-level OC}), a holonomic constraint on the acceleration with regard to the object frame can be established to compute the contact force:
\begin{equation}\label{acc_constraint}
    \v J_{oc} \ddot{\v q}+\dot{\v J}_{oc}\dot{\v q}=0
\end{equation}
Given the rigid body dynamics in Eq.~(\ref{object_arm_dynamics}), the joint acceleration and contact forces are computed as:
\begin{subequations}
    \begin{align}\nonumber
        &\ddot{\v q}=\v M^{-1}(-\v C + \v B \vg \tau + \v J_{oc}(\mathbf{q})^T \vg \lambda)\\\nonumber
        &\vg \lambda = -(\v J_{oc} \v M^{-1} \v J_{oc}^T)^{-1}(\dot{\v J}_{oc}\dot{\v q}+\alpha \v J_{oc} \dot{\v q}+\v J_{oc}\v M^{-1}\v B\vg \tau)
    \end{align}
\end{subequations}
where a restoring force $-\alpha \v J_{oc} \dot{\v q}$ is added to mitigate the numerical constraint drifting in Eq.~(\ref{acc_constraint}). The term $\v J_{oc} \v M^{-1} \v J^T_{oc}$ is referred as the inverse inertia in contact space.

\begin{figure}[t!]
    \centering
    \includegraphics[width=1\linewidth]{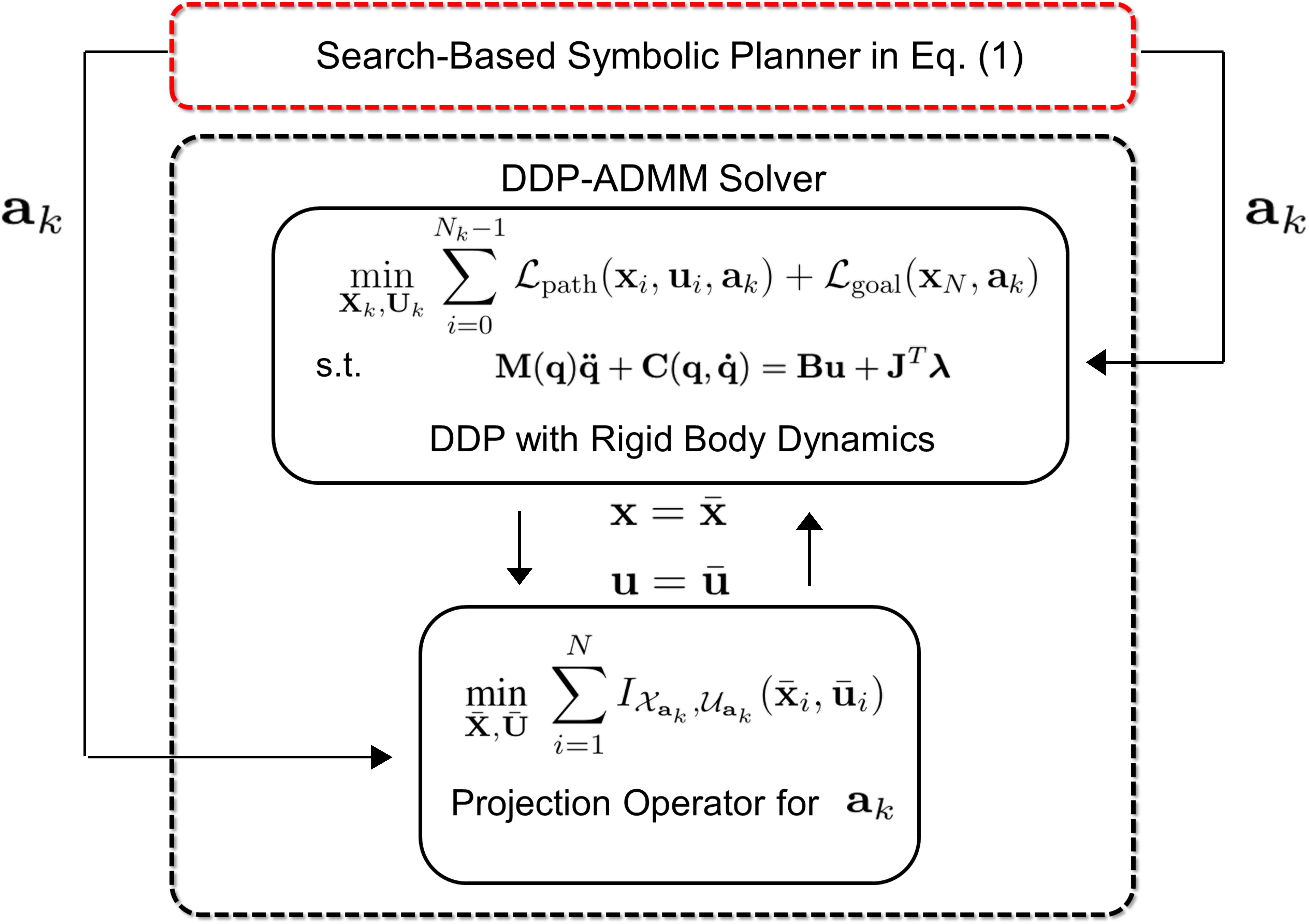}
    \caption{Bilevel optimization incorporating the high-level symbolic planner and the low-level DDP-ADMM solver. The activeness of individual projection blocks are governed by symbolic actions from the high-level search.
    }
    \label{fig:low_level_flowing}
    \vspace{-0.1in}
\end{figure}
Given the manipulator dynamics and contact constraints above, the trajectory optimization in Eq.~(\ref{low-level OC}) can be further transcribed for operator splitting
\begin{align}\nonumber
     & \underset{\vg \phi, \bar{\vg \phi}}{\text{min}} \ \sum_{i = 1}^{N} \Big[\mathcal{L}(\v x_i, \v u_i,{\v a})
        + I_{\mathcal{D}}(\v x_i, \v u_i) + I_{\mathcal{X}_{\v a},\mathcal{U}_{\v a}}(\Bar{\v x}_i, \bar{\v u}_i)\Big]\\\nonumber
        & \text{subject to} \quad \v x = \bar{\v x}, \ \v u = \bar{\v u}
\end{align}
where $\mathcal{D}=\{(\v x,\v u) \ | \ \v x_0=\v x_{\rm init}, \v x_{i+1}=\v f(\v x_i,\v u_i), i=0,1,\ldots,N-1  \}$ satisfies the dynamics constraint (\ref{RK4}). For simplicity, $\vg \phi=(\v X,\v U)$ denotes the state-control pairs, while $\bar{\vg \phi}=(\bar{\v X},\bar{\v U})$ contains all the auxiliary variables to be projected onto feasible sets.  


The trajectories are updated in a distributed manner for the $p^{\rm th}$ ADMM iteration
\begin{subequations}\label{eq:updates}
    \begin{align}
       \label{eq:primal-a} \nonumber
        \vg \phi^{p+1}=&\underset{\vg \phi}{\arg\min} \ \sum_{i = 1}^{N} \Big[\mathcal{L}(\v x_i, \v u_i,\v a)
        + I_{\mathcal{D}}(\v x_i, \v u_i) \Big]\\
         &+\frac{\rho_j}{2}\|\v x-\Bar{\v x}^{p}+\v w_j^p\|_2^2+\frac{\rho_u}{2}\|\v u-\Bar{\v u}^{p}+\v w_u^p\|_2^2 \\\label{eq:primal-b}
        \bar{\vg \phi}^{p+1}=&\Pi_{\mathcal{X}_{\v a},\mathcal{U}_{\v a}}(\v x^{p+1} + \v w_j^p,\v u^{p+1} + \v w_u^p)\\
        \v w_j^{p+1}=&\v w_j^{p}+\v x^{p+1}-\Bar{\v x}^{p+1}\\
        \v w_u^{p+1}=&\v w_u^{p}+\v u^{p+1}-\Bar{\v u}^{p+1}
    \end{align}
\end{subequations}
where $\v w_j$ and $\v w_u$ are dual variables for state constraints and torque limits, respectively. Since DDP solves unconstrained optimization efficiently, we use it to solve (\ref{eq:primal-a}). For sub-problem (\ref{eq:primal-b}), a saturation function is employed to project inputs onto admissible sets $\mathcal{X}_{\v a}$ and $\mathcal{U}_{\v a}$, separately. Therefore, the optimization problem is decomposed into an unconstrained DDP block and a projection block handling constraints. Figure \ref{fig:low_level_flowing}
demonstrates the whole framework of our operator splitting method given the high-level action $\v a_k$.

Since the DDP is used within the ADMM algorithm, it is convenient to switch between stage 1 and stage 2 (i.e. activate or deactivate all the constraint blocks) in our multi-stage search structure by simply setting the penalty parameters $\vg \rho=(\rho_j,\rho_u)$ as zero or not.
Algorithm 2 illustrates the whole process of our DDP-ADMM solver for a multi-stage search. 
For stage 1, the initial trajectory $\vg \phi^0$ is generated by a random guess for $\v u$. As for stage 2, the trajectory generated by stage 1 is employed as a warm-start $\vg \phi_{\mathcal{N}_1}^0$ for the full ADMM. Given this warm-start, each DDP step in Eq.~(\ref{eq:primal-b}) only requires very few iterations to converge (around 10 in most cases) with one ADMM iteration. The dual variables $\v w$ and trajectory for projection block $\bar{\vg \phi}$ are usually arbitrarily selected \cite{o2016conic}. Here we initialize them with zeros. The ADMM stopping criterion is designed based on the residuals of different constraints
\begin{equation}\nonumber
\begin{aligned}
    \|\v x - \bar{\v x}\|_2 \leq \epsilon_x, \quad
    \|\v u - \bar{\v u}\|_2 \leq \epsilon_u,
\end{aligned}
\end{equation}
where $\epsilon_x$ and $\epsilon_u$ are expected tolerances for separate constraints of state and control.

\section{Results}
Our SyDeBO framework is tested on a 7-DOF Kuka IIWA manipulator with a Schunk WSG gripper in simulation and hardware. The symbolic planning domain is defined in PDDL and parsed into Python using Pyperplan \cite{alkhazraji-et-al-zenodo2020}. The planning framework is implemented in C++ and Python using Drake toolbox \cite{drake}, and the source code is \href{https://github.com/GTLIDAR/tamp-manipulation}{available}\footnote{https://github.com/GTLIDAR/tamp-manipulation}. Our planner assumes fully known object geometries and locations, as perception is beyond the scope of this work. A video can be found \href{https://youtu.be/lkhr3UiiDuw}{here}\footnote{https://youtu.be/lkhr3UiiDuw}. 

\subsection{Object Sorting in Clutter}
\begin{figure}
    \centering \includegraphics[width=3.4in]{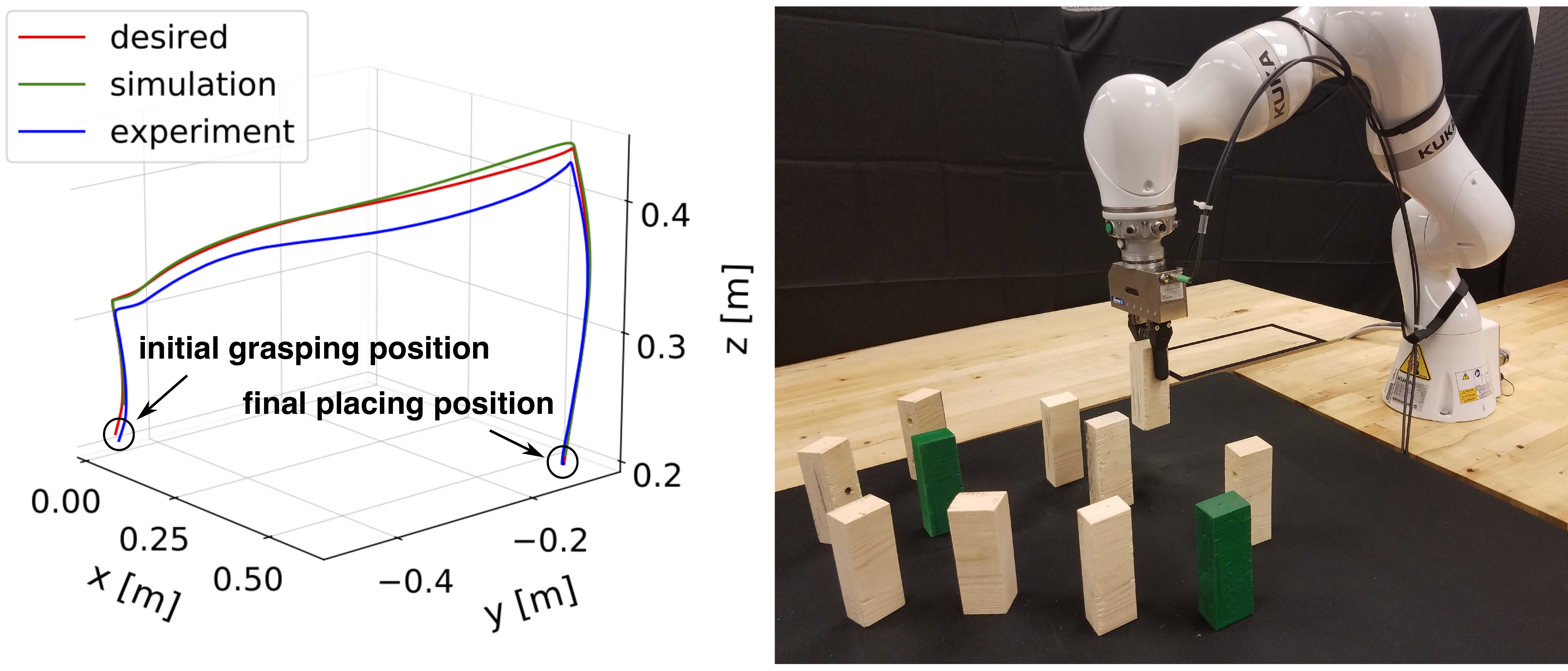}
    \caption{Control tracking performance comparison of the simulation and hardware experiment for the object sorting task in clutter.} 
    \label{fig:hardware-setup}
    \vspace{-0.1in}
\end{figure}

\begin{figure}[t]
    \centering
    \includegraphics[width=3.4in]{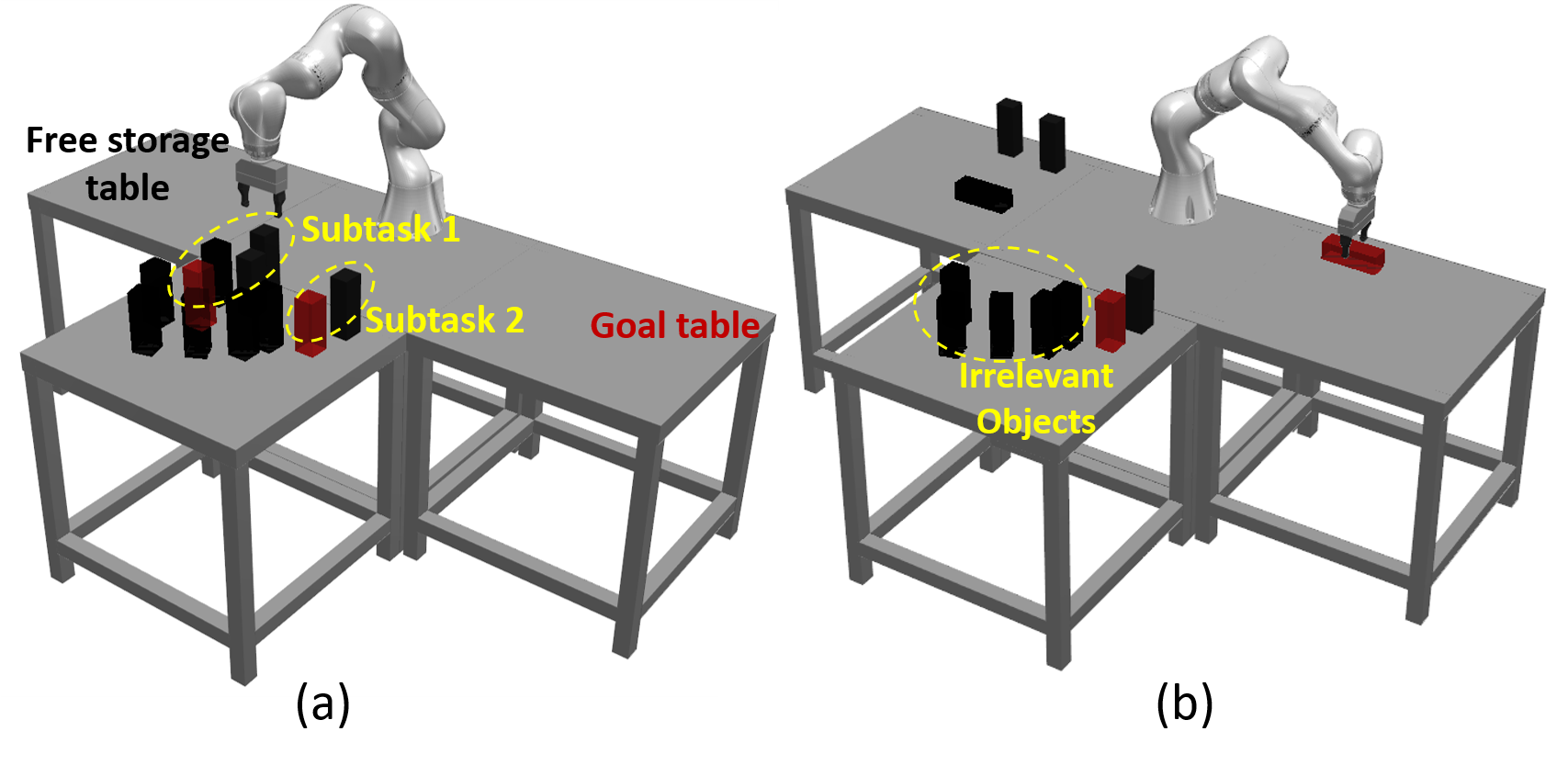}
    \caption{Object sorting in clutter. The whole object sorting problem is decomposed into two subtasks, while irrelevant objects are identified and ignored by the planner. The two subfigures shows the scenario at initial setup in (a) and after subtask 1 is completed in (b).}
    \label{fig:clutter-scenario}
    \vspace{-0.1in}
\end{figure}

The goal of this object sorting task is to move all red boxes from the cluttered area to the goal area as shown in Figure \ref{fig:clutter-scenario}.
If a black box obstructs the arm's access to a target red box, the black box needs to be moved away. In this case, the causal graph planner decomposes the planning problem into two subtasks while eliminating the irrelevant black boxes from the subtasks. 6 objects out of 11 in total are grasped. Therefore, the total planning time with the causal graph decomposition is significantly reduced. This task was evaluated in both the simulation and on the real robot hardware. The hardware setup is shown in Figure \ref{fig:hardware-setup}.

\begin{figure}
    \centering
    \includegraphics[width=1\linewidth]{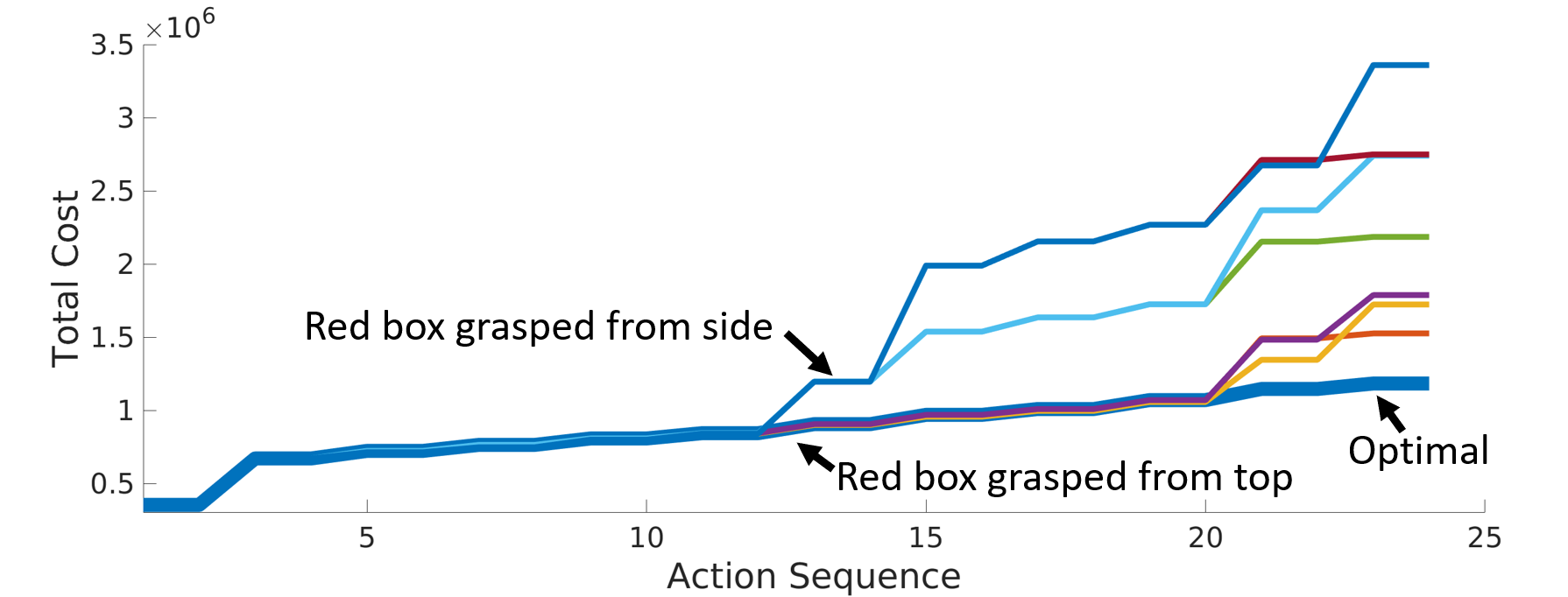}
    \caption{Accumulated total costs of diverse manipulation action sequences for the object sorting scenario in Fig.~\ref{fig:clutter-scenario}. The horizontal axis indexes the number of actions. A bifurcation appears when two different grasping actions presents. The trajectory at the bottom represents the optimal solution.}
    \label{fig:cost-vs-action}
\end{figure}

Our planner finds 8 solutions for this object sorting scenario with distinct action sequences and costs. All action sequences consist of 24 discrete actions, where the sequences and poses for manipulating the objects are defined. The selection of different action sequences results in different total costs, as shown in Figure~\ref{fig:cost-vs-action}. For example, the costs bifurcate at action 12, where the robot grasps the red box in subtask 1 from either the side or the top.

\begin{figure}[t]
    \centering
    \includegraphics[width=1.03\linewidth]{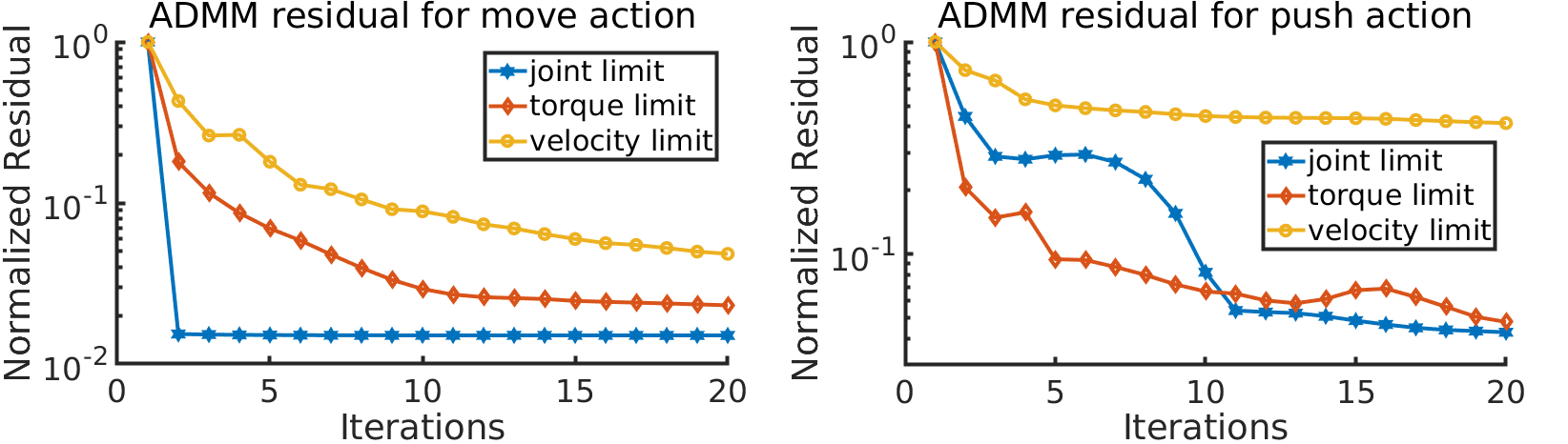}
    \caption{Normalized residuals for ADMM constraints enforced in specific actions. Contact configurations are defined specifically for each action. Although no absolute residual values are shown, all the constraints are met at a satisfactory physics accuracy. The \texttt{Push} action demonstrates a larger residual of velocity limit constraint due to our more conservative velocity limit set-up for this pushing action.}
    \label{fig:residual}
    \vspace{-0.1in}
\end{figure}

To evaluate the performance of our low-level trajectory optimizer, we show the normalized accumulated residuals along the entire trajectory with 100 knot points in Fig.~\ref{fig:residual}. The two subfigures correspond to the \texttt{Move} and \texttt{Push} actions, respectively. It is observed that in both cases, the accumulated residual for each constraint converges to high accuracy, demonstrating satisfactory constraint violations.

We evaluate the control performance of the object sorting task both in Drake simulation and the hardware in Fig.~\ref{fig:hardware-setup}. Both the simulated and the real robot have a built-in PD position controller with a servo rate of 1kHz. For a clear visualization, we only show the trajectories of a short manipulation sequence of lifting an object, moving, and placing it down (see Fig.~\ref{fig:hardware-setup}). The trajectory depicts the Cartesian end-effector position of the left fingertip. The desired and the measured hardware trajectories have an average tracking error of 1.9 cm throughout the pick-and-place motion.


\begin{figure}[t]
    \centering
    \includegraphics[width=1\linewidth]{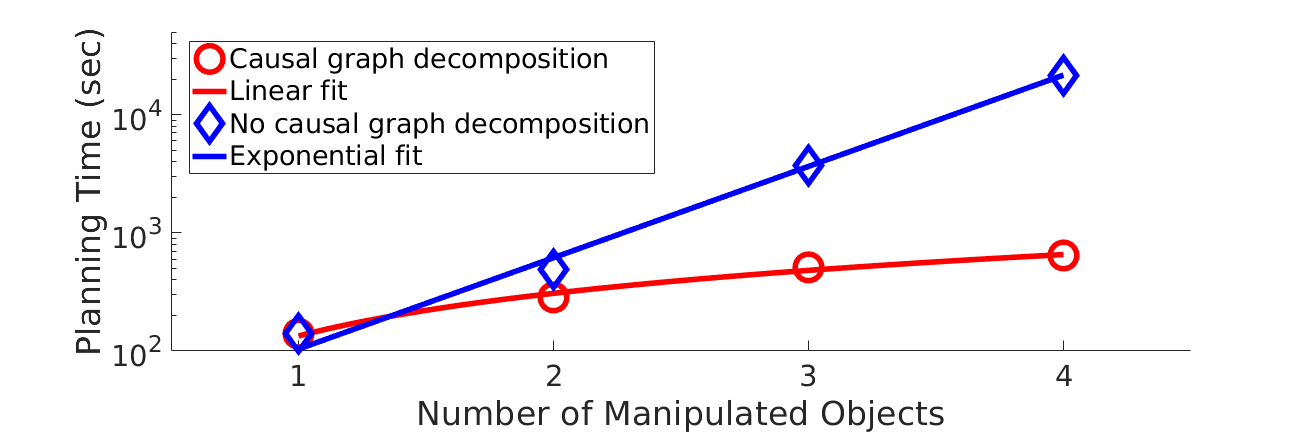}
    \caption{Total planning time in logarithmic scale w.r.t the number of objects. The results with and w/o causal graph decomposition are shown.}
    \label{fig:n-obj-vs-time-conv-belt}
\end{figure}

\subsection{Conveyor Belt Sorting}
\begin{figure}
     \centering
     \includegraphics[width=\linewidth]{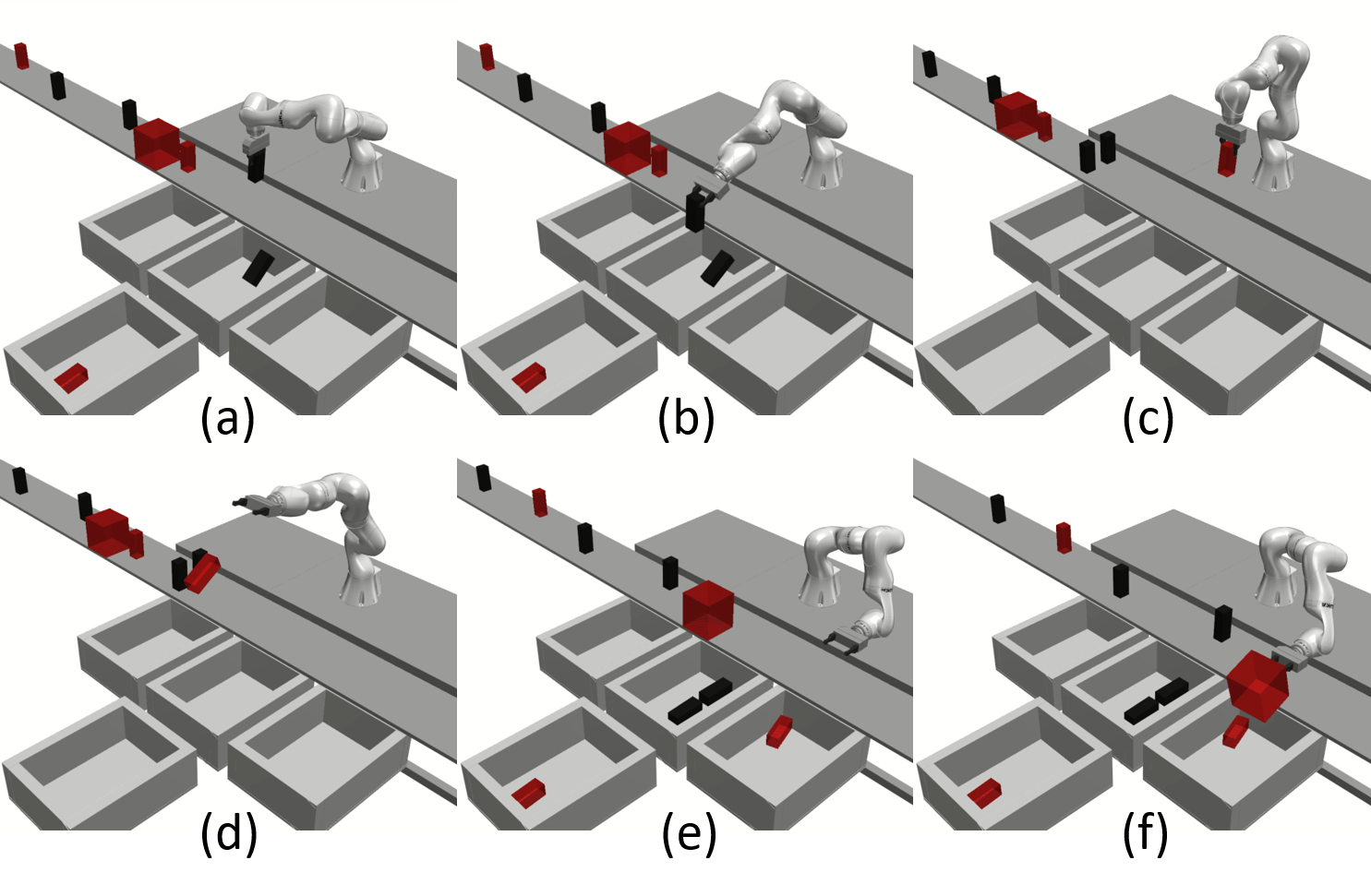}
     \caption{Snapshots of dynamic manipulation actions for object sorting on a conveyor belt. (a) and (b): \texttt{Grasp} and \texttt{Move}; (c) and (d): \texttt{Throw}; (e) and (f): \texttt{Push}. Simulation details can be checked in the attached video.}
     \label{fig:6img-pick-push-throw}
     \vspace{-0.1in}
\end{figure}

In our conveyor belt scenario, there are nine blocks cluttered on a moving conveyor belt and four bins where Bin 4 is unreachable from the robot as shown in Figure \ref{fig:scenario}. The blocks can take two different sizes (small, large), and colors (red, black). The task is to sort the blocks from the moving conveyor belt to the bins: small black blocks to Bin 2 and 3, and all red blocks to Bin 1 and 4. This problem poses logical and dynamics constraints, because the large blocks cannot be grasped, the small blocks can be grasped in different poses, and Bin 4 is unreachable from the robot. This leads to the necessity of pushing and throwing actions shown in Figure \ref{fig:6img-pick-push-throw}. 

To evaluate the scalability of our planner, we compare the planning time with and without the causal graph decomposition for sorting objects on the conveyor belt with the first four objects in Figure \ref{fig:scenario}. Except for block pairs of D2-D3 and D4-D5, the objects are fully decoupled in the conveyor belt domain since they are not blocking each other. The results in Figure \ref{fig:n-obj-vs-time-conv-belt}. shows that the total planning time grows linearly with the causal graph planner but exponentially with a single search tree. Note that the simulation result was generated by DDP without the ADMM refinementment to avoid intractable ADMM computation for the single search tree case.

In the causal graph decomposition, the size of search space does not grow exponentially with the number of manipulated objects. Instead, it depends on the coupling structure of the discrete predicates. In the extreme situation, all objects within the planning domain are decoupled, and then the total planning time with causal graph grows linearly with the number of objects. For real-world sequential manipulation, objects are often partially coupled. The causal graph decomposition will still offer computational advantages comparing to the conventional TAMP methods, depending on the level of multi-object coupling.

\section{Discussion and Conclusions}
This study proposed a TAMP framework that optimized a sequence of kinodynamically consistent plans for a diverse set of manipulation tasks over a long-horizon. This framework is generalizable to other types of manipulation skills by rapidly adding specific constraints into trajectory optimization. One of our future directions will focus on maturing our trajectory optimization method: (i) designing a collision-avoidance ADMM block by convexifying the feasible state-space \cite{schouwenaars2001mixed}, 
and (ii) applying accelerated ADMM techniques to speed up the convergence \cite{ZhouAcceleratedDynamics}. 

One limitation of our current implementation stems from the heavy computational burden of trajectory optimization for highly complex manipulation motions, in particular, this optimization solve is coupled with the exploration of a large number of symbolic nodes during discrete search. To address this computation bottleneck, our future work will develop more efficient TO algorithms through GPU-based parallel computing and Automatic Differentiation. As such, we can aim for online planning for reactive manipulation in dynamically changing environments.

Defining the symbolic planning domain requires a significant amount of human knowledge and effort. The symbolic planning domain and manipulation actions are defined by hand in PDDL, and the selection of pruning nodes in causal graph could be problem-specific. This possibly makes it challenging to apply our framework on more complex problems. Future work to enhance the flexibility of our framework includes using learning techniques to automate planning domain definition and task allocation, for example, learning compositional models for symbolic planning \cite{wang2020learning} and learning object importance for task decomposition \cite{silver2020planning}.

\bibliographystyle{IEEEtran}
\bibliography{lidar-bib.bib}

\end{document}